\documentclass[journal]{IEEEtran}
\ifCLASSINFOpdf
\else
\fi
\usepackage{amssymb}
\usepackage{subfigure}

\usepackage{amsmath,amssymb,amsfonts}
\usepackage{graphicx}
\usepackage{textcomp}
\usepackage{xcolor}
\usepackage{url}
\usepackage{amsmath,epsfig,array}
\usepackage{epstopdf}
\usepackage{multirow}
\usepackage{subfigure}
\usepackage[noend]{algpseudocode}
\usepackage{algorithmicx,algorithm}
\usepackage{fancyhdr}
\usepackage{lipsum}
\usepackage{verbatim}
\usepackage[noend]{algpseudocode}
\usepackage{algorithmicx,algorithm}
\usepackage[justification=centering]{caption}
\usepackage{float}

\begin{document}
%
\title{MRGAN360: Multi-stage Recurrent Generative Adversarial Network for 360 Degree Image Saliency Prediction}
%
%
%

\author{Pan Gao,
        Xinlang Chen,
       Rong Quan,
        and Wei Xiang
\thanks{P. Gao, X. Chen, and R. Quan are with College Of Computer Science and Technology, Nanjing University of Aeronautics and Astronautics, 
Nanjing, China.  (e-mail: pan.gao@nuaa.edu.cn; chenxinlang@nuaa.edu.cn; quanrong21@nuaa.edu.cn).}
\thanks{W. Xiang is with La Trobe University, Melbourne, Australia. (email: W.Xiang@latrobe.edu.au)}}

%
%

\markboth{Journal of \LaTeX\ Class Files,~Vol.~14, No.~8, August~2015}%
{Shell \MakeLowercase{\textit{et al.}}: Bare Demo of IEEEtran.cls for IEEE Journals}
%



\maketitle

\begin{abstract}
   Thanks to the ability of providing an immersive and interactive experience, the uptake of 360 degree image content has been rapidly growing in consumer and industrial applications. Compared to planar 2D images, saliency prediction for 360 degree images is more challenging due to their high resolutions and spherical viewing ranges. Currently, most high-performance saliency prediction models for omnidirectional images (ODIs) rely on deeper or broader  convolutional neural networks (CNNs), which benefit from CNNs’ superior feature representation capabilities while suffering from their high computational costs. In this paper, inspired by the human visual cognitive process, i.e., human being’s perception of a visual scene is always accomplished by multiple stages of analysis, we propose a novel multi-stage recurrent generative adversarial networks for ODIs dubbed MRGAN360, to predict the saliency maps stage by stage. At each stage, the prediction model takes as input the original image and the output of the previous stage and outputs a more accurate saliency map. We employ a recurrent neural network among adjacent prediction stages to model their correlations, and exploit a discriminator at the end of each stage to supervise the output saliency map. In addition, we share the weights among all the stages to obtain a lightweight architecture that is computationally cheap. Extensive experiments are conducted to demonstrate that our proposed model outperforms the state-of-the-art model in terms of both prediction accuracy and model size.
\end{abstract}


%
\IEEEpeerreviewmaketitle

\section{Introduction}
\label{sec:intro}

Recent advances in stereoscopic display technology and
graphics computing power have made virtual reality (VR)
applications in industry commercially feasible and increasingly popular \cite{TII-360,zerman2020textured,gao2019occlusion,zerman2019subjective}, which is widely used in digital entertainment, video conferencing, telemedicine and metaverse. One of the
most prominent applications for VR is displaying 360 degree image, also known as omnidirectional images (ODIs), through a head-mounted display (HMD). 360-degree VR creates an environment that surrounds the users, allowing them to look around in all directions, providing the users an immersive and interactive viewing experience. Besides VR applications, the fact that 360 degree photography  has brought a new perspective on imaging has impacted  various industries, such as real estate, retail business, science, and even the military. For example, in the online market evolution, 360 degree photograph springs up and revolutionizes the way we advertise products. Various retail industries use 360 degree image to showcase their products from the best angles, attracting countless consumers. In addition to benefiting sales, 360 degree photograph revolutionizes our perception of entertainment. Many Internet companies use 360 degree image and video in their web-player, allowing the users to engage in the social medium in a true-to-life way. 

Different from traditional 2D images, ODIs can capture scene information in the 360$^{0}$ × 180$^{0}$ field of view.
Subjects can freely view the scene in any direction. Although the applications
of ODIs are increasing rapidly, there are still some drawbacks that
limit their progress. The most notable limitation is that ODIs
have higher resolutions than traditional planar images, which makes them very difficult to store, stream or render. If we can identify
the regions that are most important or attractive to users in advance, these drawbacks can be effectively overcome. Since only a fraction of an ODI is presented in the
subject’s HMD at a given time, the subject needs to constantly move both their head and eyes to change the viewing angle, so as to complete the observation of the entire image. Thus, head and eye
fixations play an equally important role in modeling visual
attention of ODIs. Consequently, it is necessary to predict both the head and eye fixations in ODIs, which
are widely used in many realms of ODIs, such as compression,  viewport prediction, and image quality assessment \cite{zhou2021projection}. Automatically predicting regions of high saliency in a 360 degree image is useful for many applications including advertisement, customer attention marketing, and robotics for object detection. 

\begin{figure}[t]
	\centering 
	\includegraphics[width = 9.3cm]{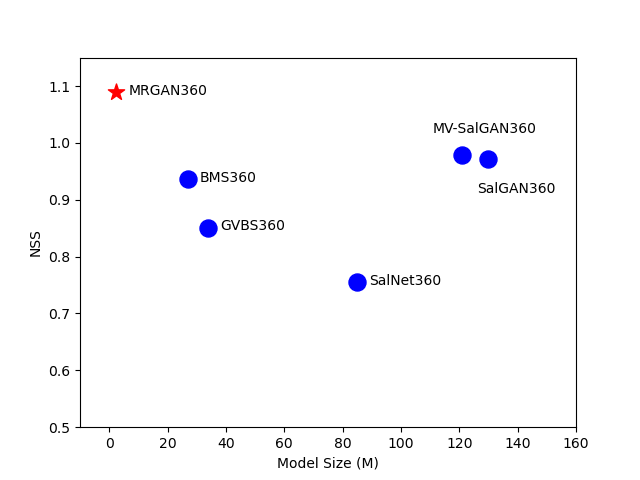}
	\caption{Comparison of the size and performance(in terms of NSS) of various state-of-the-art saliency detection models for ODIs on Salient360$!$ Grand Challenge ICME2017 \cite{rai2017dataset}. The closer the dot is to the top left corner, the better the model is.}
	\label{nssmodelsize}
\end{figure}


In the past few decades, deep convolutional neural networks (CNNs) not only brought striking performance improvements to saliency detection of 2D images \cite{TII-Saliency}, but also began to infiltrate into saliency detection of ODIs. Recently, some CNN-based ODI saliency detection methods have been proposed \cite{monroy2018salnet360, chen2021visual, chao2018salgan360, chen2020salbinet360}, and the increasingly available images/videos substantially boost their saliency detection performance. However, most CNN-based ODI saliency detection methods (\emph{e.g}., using ResNet-152 \cite{he2016identity}, Inception \cite{szegedy2016rethinking} as the backbone) achieve higher detection performance by designing deeper or broader convolutional neural networks, which will suffer from high computational costs, as shown in Fig. 1. 

\begin{figure*}[t]
	\centering
	\includegraphics[scale=0.35]{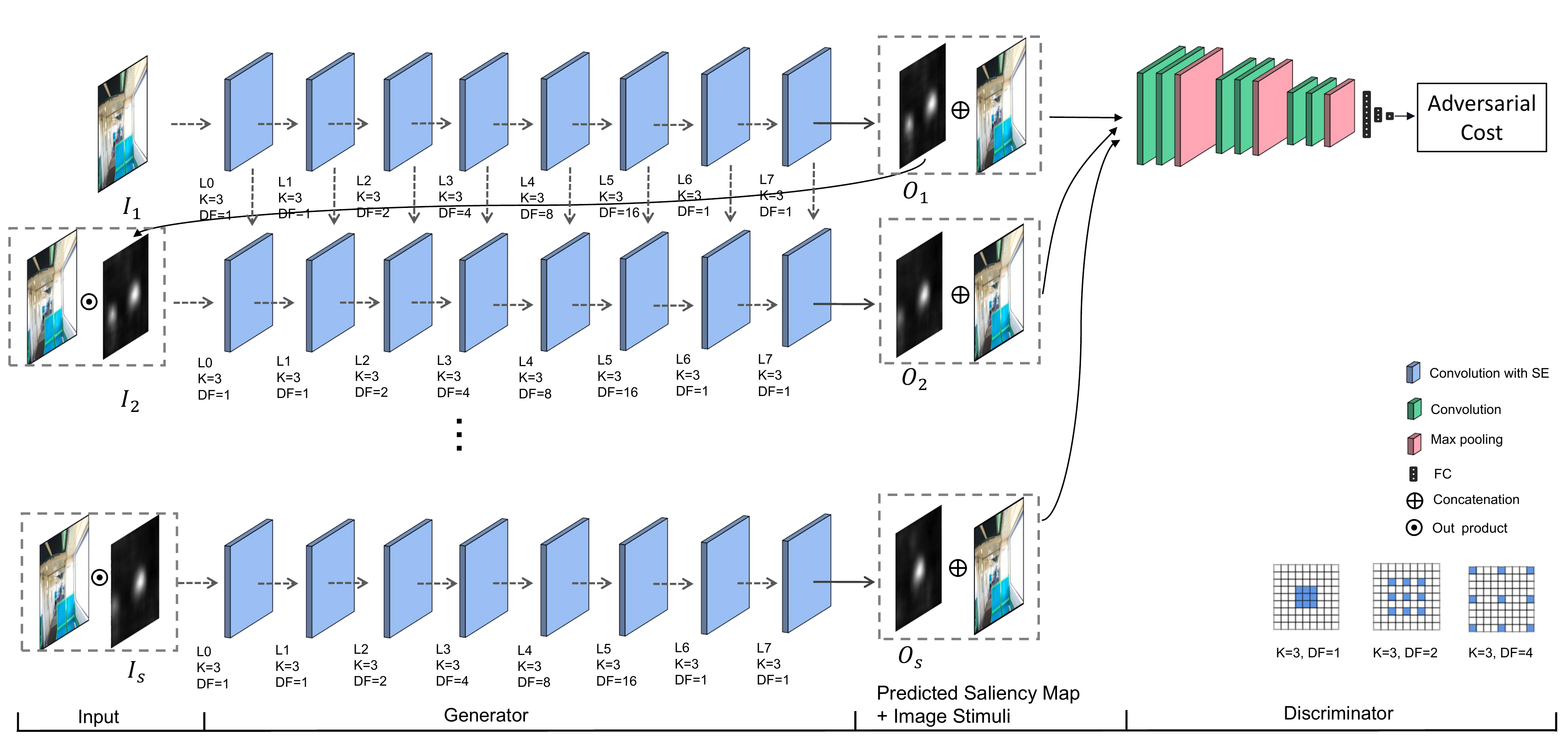}
	\caption{ The architecture of the proposed MRGAN360, where each projected view of the ODI is analyzed by multi-stage of fixation analysis. Each stage consists of a full-convolutional layer based generator and a conditional discriminator. K is the convolution kernel size,
		DF indicates the dilation factor, and S represents the number of stages.}
	\label{rgan360}
\end{figure*}

To the best of our knowledge, when free-viewing an image, humans do not analyze the entire image at a glance, they fix their gaze on one fixation of the image at a time, and move their gaze to another fixation after completing their analysis of the current fixation \cite{A06,A07}. The complete understanding of the entire image is always achieved by multiple stages of fixation analysis. Inspired by the human visual cognitive process, we propose a multi-stage saliency prediction model in this paper, which analyzes the image in a stage-by-stage manner and refines the saliency map at each stage with the goal of producing a most accurate saliency map as output. Specifically, at each stage, taking as input the original image and the output of the previous stage, we use multiple full-convolutional layers to yield a more accurate saliency map. As shown in Fig. 2, a discriminator is used at the end of each stage to give a supervision to its output saliency map. Besides, as the human vision system localizes and analyzes the current fixation based on its analysis of the previous fixations, we exploit a recurrent neural network mechanism to further model the correlations between adjacent prediction stages. We also share the network weights among all the stages to make our model lightweight, which can save computational costs to a large extent compared with the existing CNN-based models that rely on deeper and broader convolutional layers to achieve good saliency detection results, as demonstrated in Fig. 1. For brevity, the proposed Multi-stage Recurrent Generative Adversarial Networks for ODIs is termed MRGAN360. The major contributions of this paper are summarized as follows:
\begin{itemize}
    \item Instead of relying on deeper convolutional layers as in traditional approaches, we design our CNN-based saliency prediction model for ODIs by analyzing and simulating the human visual cognitive process, which has stronger interpretability and leads to a better performance.
    \item  We use a recurrent neural network mechanism to model the correlation between adjacent saliency prediction stages and share the weights among all the stages, which makes the proposed model both effective and lightweight. 
    \item  Extensive experimental results on two public benchmark datasets demonstrate that our proposed model outperforms the state-of-the-art methods in saliency prediction of ODIs.
\end{itemize}

The remainder of the paper is organized as follows. Section \ref{Sec_related_work} gives a review of saliency prediction methods for ODIs. Then, Section \ref{sec_methodlogy} describes the proposed saliency prediction framework's pipeline as well as its pre- and post-processing steps. Section \ref{sec_proposed_method} introduces the details of the proposed MRGAN360. Section \ref{sec_experimental_results} compares the MRGAN360 with other state-of-the-art models experimentally and conducts ablation experiments to analyze the effectiveness of each component of our model. Finally, Section \ref{sec_conclusion} concludes the paper and suggests some future work.

\section{RELATED WORK}\label{Sec_related_work}

In this section, we briefly review the most notable works on saliency prediction for ODIs, including traditional modeling based approaches and recent deep learning based methods. 

\subsection{Traditional Saliency Prediction Methods for ODIs}
The past decades have witnessed the tremendous success of saliency prediction for 2D images and a plethora of relevant works have emerged. However, there exist only a few methods for predicting saliency maps of ODIs. Generally, the saliency prediction methods of ODIs are developed either for eye fixation or head fixation prediction. 
Most of the existing saliency prediction methods of ODIs focus on the prediction of eye fixations. 
Specifically, Startsev \emph{et al.} \cite{startsev2018360} took projection distortions, equator bias and vertical border effects
into account and proposed a new saliency prediction method for predicting eye fixations. 
Battisti \emph{et al.} \cite{battisti2018feature} developed a saliency prediction model by utilizing low-level and high-level image features. 
Regarding the prediction of head fixations for ODIs, only a few saliency prediction works \cite{lebreton2018gbvs360, zhu2018prediction} have been proposed. 
For instance, Lebreton \emph{et al.} \cite{lebreton2018gbvs360} extended traditional 2D saliency prediction models to omnidirectional scenes, and
developed two new models for predicting head fixations for ODIs, namely GBVS360 and BMS360. 
Besides, Zhu \emph{et al.} \cite{zhu2018prediction} proposed a multiview projection approach for predicting the saliency maps of head fixations on ODIs. These two
models are built upon the traditional 2D images saliency prediction models, i.e., Boolean Map based Saliency model (BMS) \cite{zhang2015exploiting} and Graph-Based Visual Saliency (GBVS) \cite{harel2007graph}.

\subsection{CNNs-based Saliency Prediction Methods for ODIs}
More recently, the performance of saliency prediction has seen considerable improvements due to the rapid advance of CNNs. The most representative works for CNNs-based saliency prediction for ODIs are \cite{monroy2018salnet360, chen2021visual, chao2018salgan360, chen2020salbinet360}. 
In particular, Monroy \emph{et al.} \cite{monroy2018salnet360}  proposed a new saliency prediction method, which incorporated spherical coordinates for feature learning and fine-tuned traditional 2D CNN models for eye fixations prediction for ODIs. 
Chen \emph{et al.} \cite{chen2021visual} presented a new loss function in consideration of both distribution- and location-based evaluation indicators, which is then introduced to fine-tune the network for saliency prediction for ODIs. 
In \cite{pan2017salgan}, inspired by generative adversarial networks (GANs), Pan \emph{et al.} adopted the adversarial training mechanism of GANs and proposed  to predict the saliency of head fixations on normal images.
Chao \emph{et al.} proposed to extend SalGAN~\cite{pan2017salgan} to SalGAN360~\cite{chao2018salgan360} by fine tuning the SalGAN with a different loss function to predict both global and local saliency maps.
Later, Chen \emph{et al.} proposed a local-global bifurcated deep network for saliency prediction for 360° images, which is dubbed SalBiNet360 \cite{chen2020salbinet360}. On the basis of SalGAN360, Chao \cite{chao2020multi} proposed a method called MV-SalGAN360 to estimate the 360$^{o}$ saliency map which extracts salient features from the entire 360$^{o}$ image in each viewport with three different Field of Views (FoVs).

However, all the aforementioned methods depend on deeper and wider CNNs to achieve better results, thus incurring huge computational costs. Different from these computationally expensive methods, we propose a novel lightweight model for predicting the eye fixations on ODIs, where we predict saliency maps of ODIs in a multi-stage manner by simulating the human visual perception process. We introduce recurrent neural networks into our proposed model to model the correlations between the prediction stages and share weights among all the prediction stages to make the proposed model both accurate and lightweight.

\section{METHOD}
\label{sec_methodlogy}
\begin{figure}[t]
	\centering 
	\includegraphics[width = 10cm]{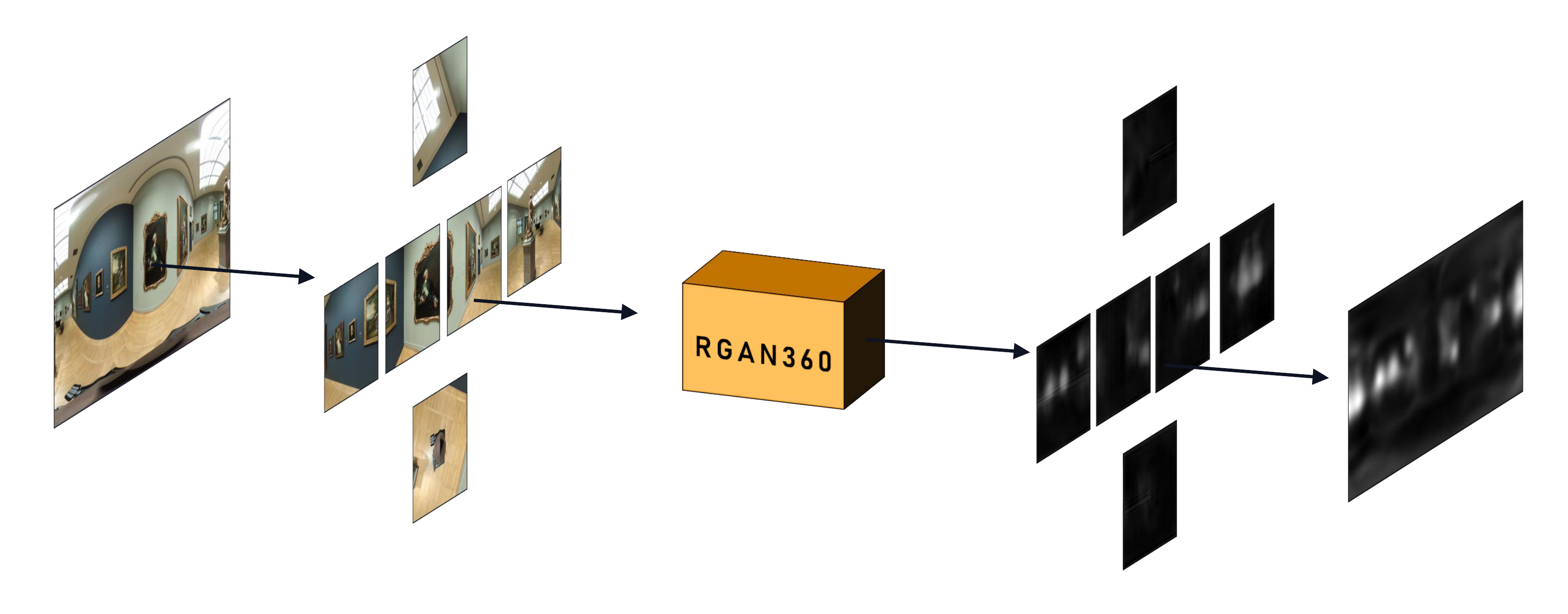}
	\caption{ODI saliency prediction pipeline.}
	\label{pipeline}
\end{figure}

In this section we detail the pipeline used to obtain the saliency map for an input ODI. Fig.~\ref{pipeline}  shows the complete pipeline of the proposed framework.
The input ODI is first projected into six rectilinear images using
the pre-processing steps described in Section~\ref{sec_preprocessing}. Then, each of these rectilinear images is
fed into the MRGAN360 to obtain a saliency map. The details of the MRGAN360 will be illustrated in Section \ref{sec_proposed_method}.
Finally, the output saliency maps for all the rectilinear images are combined into a single saliency map using the
post-processing technique described in Section \ref{Section:Post}.

\subsection{Pre-processing}\label{sec_preprocessing}
\begin{figure}[t]
	\centering 
	\includegraphics[width = 9cm]{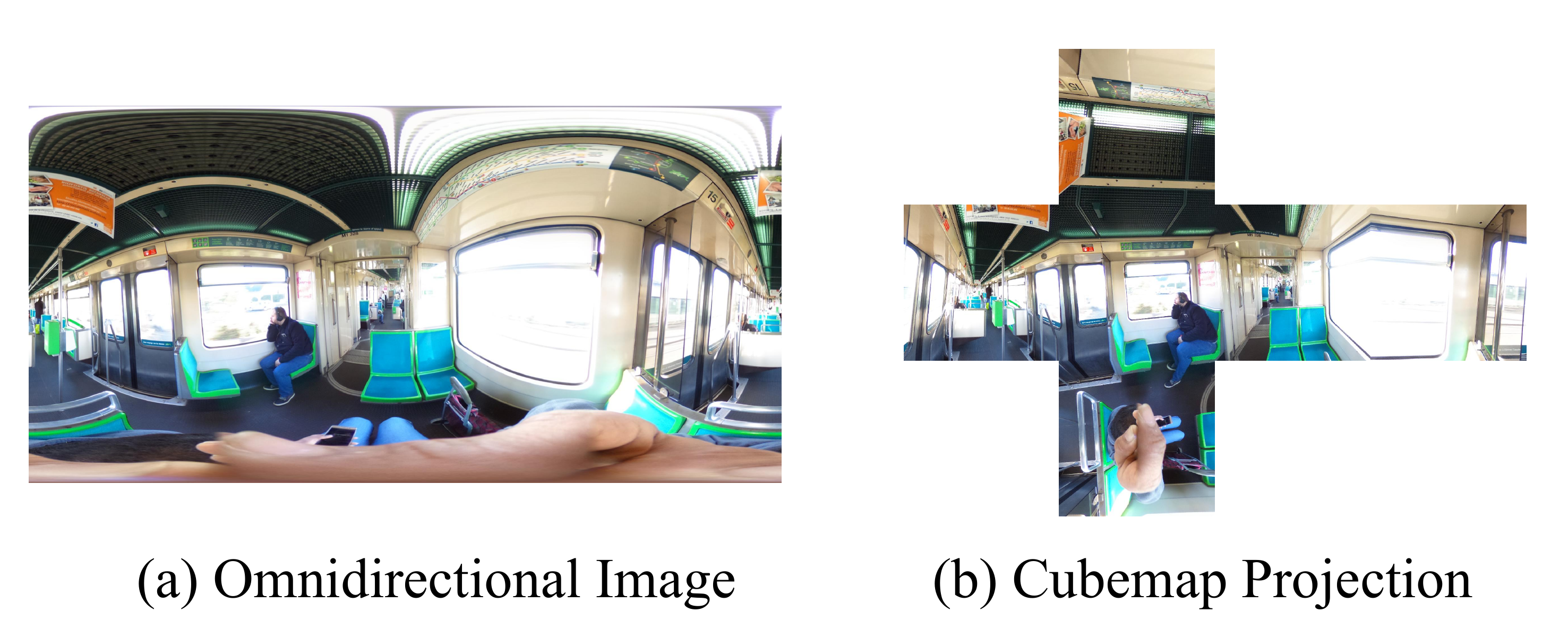}
	\caption{Examples of the cubemap projection for omnidirectional content.}
	\label{cmp}
\end{figure}

For a 360$^{o}$ image, mapping from a spherical surface to a flat plane would induce stretching distortion, and this phenomenon is more pronounced at the two poles \cite{gao2020quality}. Fig.~\ref{cmp} shows an example of how to project a full ODI into rectilinear images that users actually observe by using an HMD. As can be seen from this figure, when we view this ODI directly, it is difficult to identify the south pole in Fig.~\ref{cmp}(a) as a whole handle due to distortion. To reduce the impact of this kind of distortion on saliency prediction, we need to render the viewing cone by rendering the field of view (FOV). As shown in Fig. \ref{cmp}, a popular projection method is cubic mapping, which performs rectilinear projection on 6 faces of the cube
with 90$^{o}$ FOV each. In this paper, we use cubic mapping to project an input ODI into six rectilinear images.

\subsection{Post-processing}\label{Section:Post}
The proposed MRGAN360 takes as input each of the six rectilinear images and yield six saliency maps. Then, the generated saliency maps are combined into a final complete saliency map for the input ODI. In each cube face, the distortion is not as obvious as in the equirectangular image.
But there will be some visual artifacts caused by intensity and structure misalignment when combing these saliency maps.   
Therefore, we utilize dense cubic projection, which samples viewports in
every 10$^{o}$ in the longitude and latitude, then back-projects viewport saliency maps into equirectangular format, and averages the equirectangular maps to output a final saliency map. In this way, the undesirable borderlines due to discontinuity between the cube faces can be removed by averaging the viewpoints of different angles.

\section{Proposed Saliency Prediction Model}\label{sec_proposed_method}
We propose a multi-stage recurrent generative adversarial network for image saliency prediction for each rectilinear image. The framework of our proposed network is presented in Fig.~\ref{rgan360}. We predict saliency maps stage by stage. At each
stage, we use the context aggregation network with Squeeze-and-Excitation (SE) blocks \cite{hu2018squeeze}  to predict saliency maps. Dilated convolution is used in our network to enlarge the receptive field
and acquire more contextual information. By using SE blocks, we can assign
different alpha values to various feature maps according to their interdependencies in each convolution layer. As saliency maps are predicted in multiple stages, useful
information for saliency prediction in preceding stages can guide the learning in later
stages. So we incorporate the RNN architecture with a memory unit to make full use of the information learnt in preceding stages.

\subsection{Baseline Model}
As shown in the first row of Fig.~\ref{rgan360}, the backbone model of our proposed MRGAN360 is a forward network without recurrence. Essentially this is a model similar to a conditional generative adversarial network. The generator is implemented by extending the Context Aggregation Net (CAN) \cite{chen2017fast} with an SE block, which is a full-convolutional network. As shown in
Fig.~\ref{rgan360}, we employ 8 convolutional layers with the SE for one stage. Since a large receptive field is very helpful for acquiring richer contextual information, we use the dilated convolution in our network. For
layers L1 to L5, the dilation rate increases exponentially from 1 to 16, which leads to
an exponential growth of the receptive field of every element. As we treat the first
layer as an encoder to transform an image to feature maps, and the last two
layers as a decoder to map reversely, we do not apply dilation for layers L0, L6
and L7. Moreover, we use $3 \times 3$ kernel convolution  {and 24 channels} from layers L1 to L6. To recover the
RGB channels of a color image or the gray channel for a gray scale image, we adopt
the  $1 \times 1$ convolution for the last layer L7. Every convolution except the
last one is followed by a nonlinear operation. 

The one-channel saliency map generated by the generator and the original three-channel input are cascaded into four-channel data, which is then used as input to the discriminator. {In detail, the discriminator is comprised of $3\times3$ and $1\times1$
kernel convolutions interspersed with three pooling layers,
 followed by three fully-connected layers. The details of the proposed discriminator network are shown in Table 1.
}The convolutional
layers all use Leaky ReLU  with $\alpha=0.2$ activation, while the fully-connected layers
employ Tanh activation, with the exception of the final layer, which adopts Sigmoid activation.

\begin{table}[th]
	\centering
	\small
	\caption{Architecture details of the proposed discriminator network.}
	\label{tab:my-table}
	\begin{tabular}{lrrrrr}
        \hline layer & depth & kernel & stride & pad & activation \\
        \hline 
        conv1 & 3 & $1 \times 1$ & 1 & 1 & ReLU \\
        conv2 & 32 & $3 \times 3$ & 1 & 1 & ReLU \\
        pool1 & & $2 \times 2$ & 2 & 0 & $-$ \\
        \hline 
        conv3 & 64 & $3 \times 3$ & 1 & 1 & ReLU \\
        conv4 & 64 & $3 \times 3$ & 1 & 1 & ReLU \\
        pool2 & & $2 \times 2$ & 2 & 0 & $-$ \\
        \hline 
        conv5 & 64 & $3 \times 3$ & 1 & 1 & ReLU \\
        conv6 & 64 & $3 \times 3$ & 1 & 1 & ReLU \\
        pool3 & & $2 \times 2$ & 2 & 0 & $-$ \\
        \hline fc1 & 100 & $-$ & $-$ & $-$ & tanh \\
        fc2 & 2 & $-$ & $-$ & $-$ & tanh \\
        fc3 & 1 & $-$ & $-$ & $-$ & sigmoid \\
        \hline
    \end{tabular}
\end{table}

\subsection{Recurrent Model}
 In the MRGAN360, we use recurrent neural networks across multiple stages to model the correlations among the prediction stages. Specifically, as shown in Fig. 2, we use the neural units of the preceding prediction stage to guide the generation of neural units in the next prediction stage. This process can be formulated as
\begin{equation}
\mathbf{O}_{s} =f_{C N N}\left(\mathbf{I}_{s}\right), 1 \leq s \leq S
\end{equation}
\begin{equation}
\mathbf{I}_{s+1} =\mathbf{I}_{1} \odot \mathbf{O}_{s}, 1 \leq s \leq S
\end{equation}
where $\mathbf{I}_s$ is the input of the $\emph{s}$-th stage, $S$ is the total number of stages, and $\mathbf{O}_{s}$ is the output of the \emph{s}-th stage. $\mathbf{O}_{s}$ is the predicted saliency map with one channel, and $\mathbf{I}_{s} (s>1)$ is the combination of the original image and the predicted saliency map with three channels. For the generation of $\mathbf{I}_{s+1} (s>1)$, the predicted map $\mathbf{O}_{s}$ is applied to input $\mathbf{I}_{1}$ via an element-wise product with each channel of the input. As these different stages work together to refine predicted saliency regions, the input images in stages $\left\{\mathbf{I}_{1}, \mathbf{I}_{2}, \cdots, \mathbf{I}_{s}\right\}$ can be regarded as a temporal sequence of saliency image predictions. It is thus more meaningful to investigate the recurrent connections between features of different stages rather
than only using the recurrent structure. So we incorporate the recurrent neural network (RNN) with memory unit to make better use of information in preceding stages and guide feature learning in later stages.
As shown in Fig.~\ref{rgan360}, we use the dot-product of the output image of the preceding stage and the original image as the input of the current stage, and consider the feature connections between these stages. At the discriminator, the predicted saliency map from the generator concatenated with the original image is used as input.

Compared with the convolutional unit, the ConvRNN, ConvGRU and ConvLSTM units have two, three and four times parameters, respectively. Considering the number of parameters and performance, we choose the ConvGRU as the backbone of our proposed architecture. The Gated Recurrent Units (GRU) is a very popular recurrent unit in sequential models. Its convolutional version ConvGRU is adopted in our model. Denote by $x_{s}^{j}$ the feature map of the $j$th layer at the $s$th stage, which can be computed based on $x_{s}^{j-1}$ (feature map in the preceding layer of the sames stage) and $x_{s-1}^{j}$ (feature map in the same layer of the preceding stage)
 \begin{align}
 z_{s}^{j}  &= \sigma\left(W_{z}^{j} \circledast x_{s}^{j-1}+U_{z}^{j} \circledast x_{s-1}^{j}+b_{z}^{j}\right) & \\
 r_{s}^{j}  &= \sigma\left(W_{r}^{j} \circledast x_{s}^{j-1}+U_{r}^{j} \circledast x_{s-1}^{j}+b_{r}^{j}\right)  &\\
 n_{s}^{j}  &= \tanh \left(W_{n}^{j} \circledast x_{s}^{j-1}+U_{n}^{j} \circledast\left(r_{s}^{j} \odot x_{s-1}^{j}\right)+b_{n}^{j}\right) & \\
 x_{s}^{j}  &= \left(1-z_{s}^{j}\right) \odot x_{s-1}^{j}+z_{s}^{j} \odot n_{s}^{j} & \label{eq2}
\end{align}   
where $\sigma$ is the Sigmoid function $\sigma(x)=1 /(1+\exp (-x))$, $\odot$ denotes element-wise multiplication, and $\circledast$ denotes convolution. 
$W$ is a dilated convolutional kernel, and $U$ is a conventional kernel of size $3\times3$ or $1\times1$.

\subsection{Loss Function}
As a discriminator is used in each stage to refine the prediction results, the loss function of our model contains the content loss and the adversarial loss. The content loss is computed on a per-pixel basis, where each pixel
value of the predicted saliency map is compared with its corresponding peer from the ground truth map.
For the baseline model in the recurrent neural network, we define the content loss function as follows
\begin{equation}
\begin{split}
L_1\left({y}, \tilde{{y}}, {y}^{f i x}\right)=
&L_{K L d i v}\left({y}, \tilde{y}\right) - L_{C C}\left({y}, \tilde{y}\right) \\
&- L_{N S S}\left(\tilde{y}, {y}^{f i x}\right)
\end{split}           
\end{equation}
where $\tilde y$, ${y}$ and $y^{f i x}$ are the predicted saliency map, the ground truth density distribution, and the groundtruth binary fixation map, respectively. The binary fixation map shows discrete fixation locations, while the ground truth density distribution regards the human fixation on the image as a continuous probabilistic distribution, which is usually converted from the discrete fixations by Gaussian smoothing. In (7), $L_1$ combines three evaluation metrics, namely the Kullback-Leibler Divergence (KL-Div), the Linear Correlation Coefficient (CC), and the Normalized Scanpath Saliency (NSS). The content loss is used to pretrain each stage of the proposed model without adversarial learning. 

The adversarial training for the saliency prediction problem has some important differences from conventional generative adversarial networks. First, the objective is to fit a deterministic function that predicts realistic saliency values from input
images, rather than realistic images from random noise. As such, in our case the input to the generator (saliency prediction
network) is not random noise but an image. Second, the inclusion of the input image to the discriminator that a saliency map corresponds to is essential, because our goal is to not only have the two saliency maps becoming indistinguishable but also satisfy the condition that they both correspond to the same input image.
We thus concatenate both the
image and its saliency map as inputs to the discriminator network. Finally, when using generative adversarial networks to generate
realistic images, there is generally no ground truth to compare
against. In our case, however, the corresponding ground truth
saliency map is available. The final loss
function for the discriminator of the saliency prediction network during adversarial
training can be formulated as
\begin{equation}
\begin{aligned}
\mathcal{L}_{c G A N}(G, D)=& \mathbb{E}_{x, y}[\log D(x, y)]+\\
& \mathbb{E}_{x}[\log (1-D(x, G(x)))]
\end{aligned}
\end{equation}
where an adversary $D$ tries to maximize the objective function $D^{*}=\arg\max _{D} \mathcal{L}_{c G A N}(G, D)$, i.e., trying to distinguish the predicted saliency map from the ground truth saliency. $x$ represents the image stimuli, i.e., $\mathbf{I}_{1}$. $D(x, G(x))$ denotes the probability that the generated saliency map $G(x)$ is the ground truth. $D(x, y)$ is the probability that $y$ is classified as the ground truth.

With adversarial training, the loss function for the generator is written as follows
\begin{equation}
\mathcal{L}_{c G A N}(G)=\mathbb{E}_{x}\left[\log (1-D(x, G(x)))+L_1\left({y}, G(x), {y}^{f i x}\right)\right]
\end{equation}
where the generator learns the mapping $G:\{x\} \rightarrow y$ from observed image $x$ to saliency $y$, and  $G$ tries to minimize this objective function, i.e., $G^{*}=\arg\min _{G} \mathcal{L}_{c G A N}(G)$.

\section{Experimental Results}
\label{sec_experimental_results}
\subsection{Evaluation Datasets and Metrics}
The proposed model is evaluated on two publicly available
datasets, i.e., Salient360$!$ Grand Challenge ICME2017 \cite{rai2017dataset} and
Saliency in VR \cite{sitzmann2018saliency}. Salient360$!$ consists of 65 omnidirectional images, for which 
observers’ head and eye movements were recorded. 40 images are 
for training and 25 for testing. This dataset is jointly realized by the University of Nantes and Technicolor, and co-sponsored by  Oculus, ITN MCSA PROVISION, and ``Quest Industries Creatives" a Research, Education, and Innovation cluster in France \cite{rai2017dataset}.
Saliency in VR consists of 22 images, 
for which observers’ head movements were recorded. This dataset is created by capturing and analyzing gaze and head orientation data of 169 users exploring stereoscopic and static omni-directional panoramas for a total of 1980 head and gaze trajectories. 
The images in both datasets consist of indoor and outdoor scenes, such as industrial architecture, interior living room, art gallery, landscape, and so on.
The performance of the proposed MRGAN360 is evaluated using four metrics 
commonly used in saliency prediction, namely Kullback-Leibler divergence (KL), Linear Correlation coefficient (CC), Normalized 
Scanpath Saliency (NSS), and the Area Under the receiver operating characteristic Curve (AUC). KL and CC evaluate the 
saliency density distribution between two saliency maps, while 
the other two metrics evaluate the location distribution of eye 
movement between the saliency map and the binary fixation map. 
More details about the four metrics can be found in \cite{bylinskii2018different}. The 
performance of dataset Salient360$!$ is evaluated using KL, CC, NSS, and 
AUC, while for the Saliency in VR dataset, only CC is employed to 
evaluate the performance as stated in \cite{sitzmann2018saliency}. The best result of each 
experiment is shown in bold in the following subsections.

\subsection{Training}
In the absence of a large dataset of 360° images, the MRGAN360 has to be pre-trained on SALICON \cite{jiang2015salicon}. The dataset contains 15,000 2D images and their corresponding saliency maps, of which 10,000 are utilized for training and 5,000 for validation.
The 2D images are input to the model and predicted 2D saliency maps are generated. Pre-training does not include discriminators and all the images and saliency maps are scaled to a resolution of 256 × 192. After pre-training the model on SALICON, the dataset used to fine tune the network contains 40 images
of head and eye movements provided by the University of Nantes \cite{rai2017dataset}. We use 30 images to train and 10 images to validate our model. In the multiple cube mapping step, for each training and validation image, we rotate the cube every 30$^{o}$, i.e., 0$^{o}$, 30$^{o}$, 60$^{o}$ both horizontally and vertically (note that rotating 90$^{o}$ equals 0$^{o}$). There are thus 3 × 3 = 9 rotations to obtain 9 × 6 = 54 rectilinear images for each image. Totally, we produced 30 × 54 = 1620 training samples and 10 × 54 = 540 validation samples. Note that although a smaller rotation angle produces more rectilinear images in different viewpoint, it also causes more overlaps between the rectilinear images which may give rise to overfitting in our model. 
In fine tuning, we add the discriminator and begin adversarial training and the number of stage equals 6. The input to the discriminator network is RGB+Saliency images of size 256 × 192 × 4 containing both the source
image channels and (predicted or ground truth) saliency. We fine tune the network model on the 1620 rectilinear images for 100 epochs using the adaptive moment estimation method (Adam) with a batch size of six and a learning rate of $5 \times 10^{-6}$. During the adversarial training, we alternate the training of the saliency prediction network and discriminator network after each iteration (batch). To avoid overfitting, we additionally use standard weight decay as regularization.

\subsection{Ablation study on adversarial learning}
In order to validate the improvements brought about by our proposed adversarial training method, we define two scenarios for comparison. In the first scenario we use only the generator, i.e., the recurrent network without discriminators, while in the second scenario we use the whole proposed model. The experimental results are shown in Table~\ref{Tabel:ablation}, which lists the average KL, CC, NSS, and AUC for the 25 test images in Salient360$!$ Grand Challenge ICME2017. As can be seen from the table, the results are clearly improved considerably in all the metrics after adding the discriminator for each stage.
\begin{table}[h]
	\small
	\centering
	\caption{Performance comparison after averaging the results of the test frames in two experimental scenarios on the Salient360$!$ dataset \cite{rai2017dataset}.}
	\label{tab:my-table}
	\renewcommand\arraystretch{1.1}
	\begin{tabular}{|c|c|c|c|c|}
		\hline
		Method        & KL$\downarrow$  & CC$\uparrow$  & NSS$\uparrow$   & AUC$\uparrow$  
		\\
		\hline
		w/o discriminators     & 0.49    & 0.583     & 0.950      & 0.744         
		\\ 
		MRGAN360 & \textbf{0.401} & \textbf{0.658} & \textbf{1.09} & \textbf{0.784} 
		\\
		\hline
	\end{tabular}
	\label{Tabel:ablation}
\end{table}

\subsection{Ablation Study on Recurrent Refinement}
\begin{figure}[ht]
	\centering
	\includegraphics[scale=0.27]{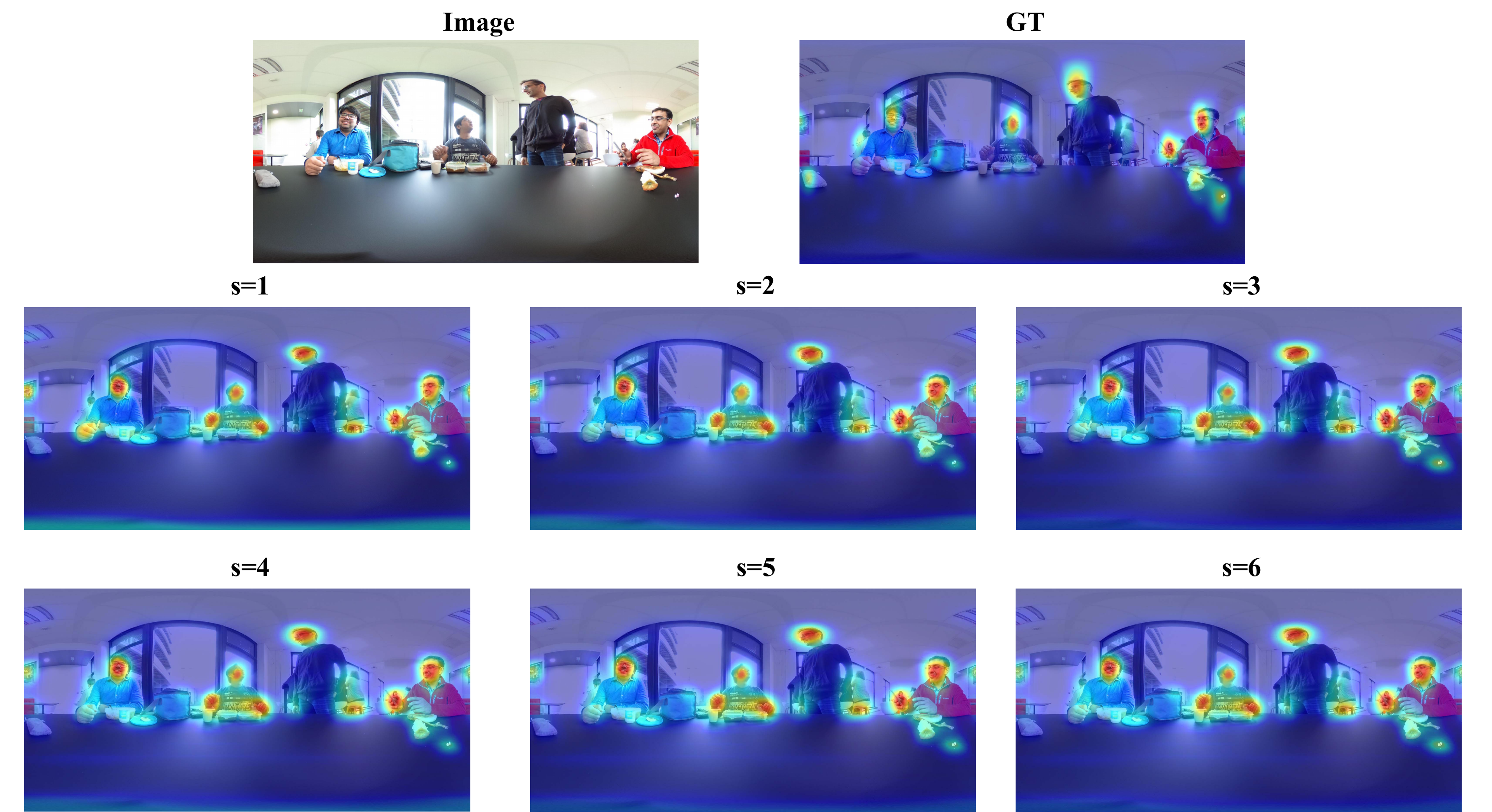}
	\caption{Progressive refinement of predictions performed by learning via  multiple stages.}
	\label{progressive_refinement}
\end{figure}

We also validate the effect of learning via multiple stages for the proposed model. Fig.~\ref{progressive_refinement} shows saliency predictions at different stages, using the output of a preceding stage as input to the next stage. The second and  third rows of Fig.~\ref{progressive_refinement} show a progressive change of focus
in the estimated saliency map, which is, regions which were wrongly predicted as salient
are progressively changed to non-salient, and truly salient regions are correctly identified. This refinement results in a
significant enhancement of the predictions. The corresponding objective results measured by various metrics are presented in  Table~\ref{Table:RefinementPerformanceSalient360}. As can be seen from the table, the prediction result is gradually improved, and
progressively close  to the ground truth. 
We also test using more stages, but found that the performance gains become saturated after six stages.
\begin{table}[th]
	\centering
	\small
	\caption{Resluts on Salient360! dataset \cite{rai2017dataset} in which each stimuli is employed as the input of the rest of MRGAN360 at different stages.}
	\label{tab:my-table}
	\renewcommand\arraystretch{1.1}
	\begin{tabular}{|c|c|c|c|c|c|}
		\hline Methods & KL$\downarrow$ & CC$\uparrow$ & NSS$\uparrow$ & AUC$\uparrow$ \\
		\hline 
		s=1 & $0.44$  & $0.648$ & $0.921$ & $0.776$ \\
		s=2 & $0.428$ & $0.648$ & $0.938$ & $0.779$ \\
		s=3 & $0.435$ & $0.638$ & $0.95$ & $0.779$ \\
		s=4 & $0.427$ & $0.651$ & $1.016$ & $0.782$ \\
		s=5 & $0.408$ & $0.656$ & $1.05$ & $0.782$ \\
		s=6 & $\textbf{0.401}$ & $\textbf{0.658}$ & $\textbf{1.09}$ & $\textbf{0.784}$ \\
		\hline
	\end{tabular}
	\label{Table:RefinementPerformanceSalient360}
\end{table}

\subsection{Comparison with State-of-the-Art Models}
\begin{figure*}[h]
	\centering
	\includegraphics[scale=0.15]{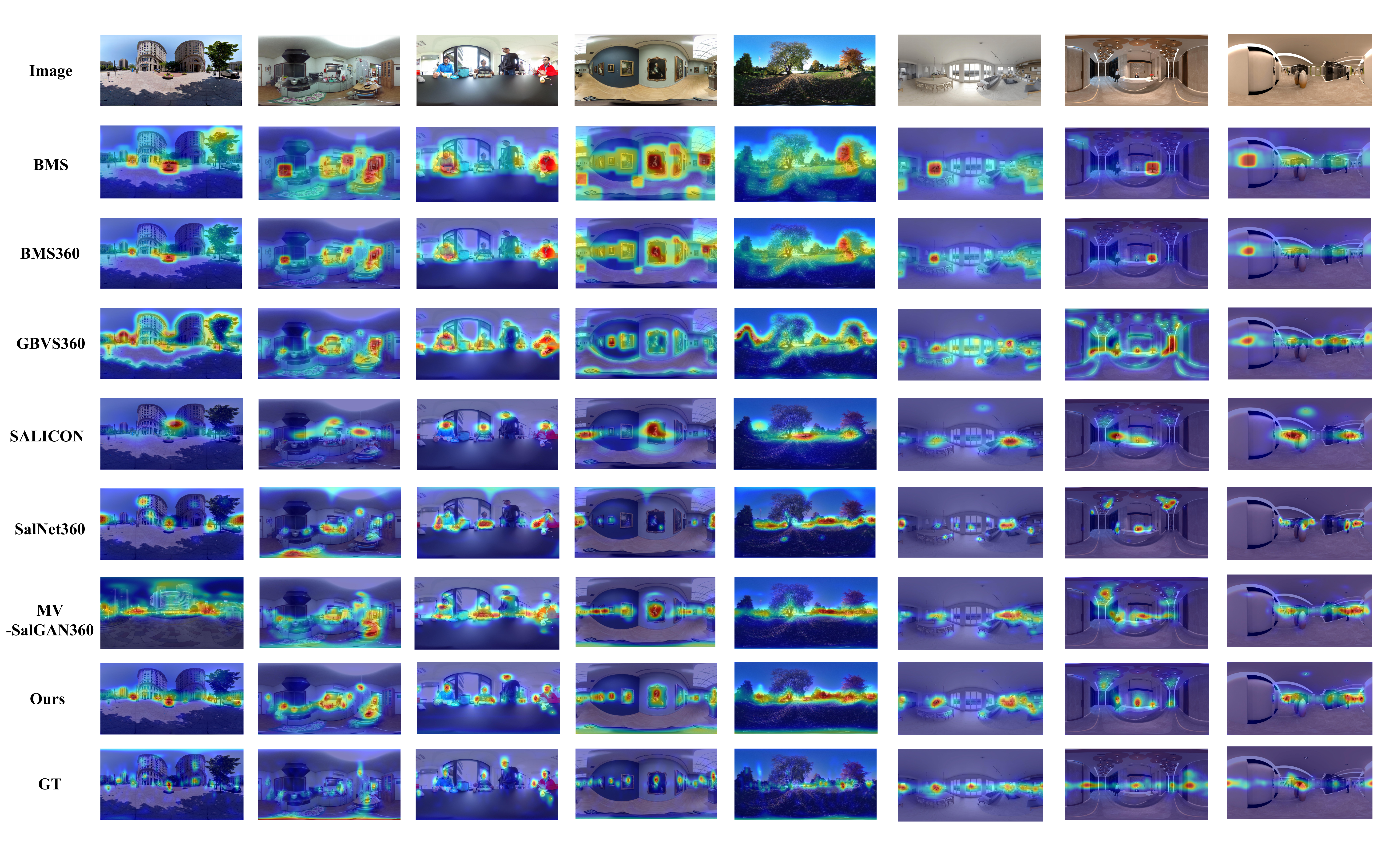}
	\caption{Visual comparison of our results with other approaches for predicting saliency maps of Head+Eyes on the Salient360$!$ dataset \cite{rai2017dataset} and
		the Saliency in VR dataset. The first five columns are ODIs from the Salient360$!$ dataset, and the last three columns are ODIs from the Saliency in VR dataset \cite{sitzmann2018saliency}.}
	\label{comparsion}
\end{figure*}

For the sake of performance comparison, all the methods are performed on the same test dataset. The results of our model and other models in terms of these metrics can be seen in Tables \ref{Table:PerformanceSalient360} and \ref{Table:performanceVR}. Overall, it is easy to see that our model outstrips the other state-of-the-art methods in terms of the prediction performance. Table~\ref{Table:PerformanceSalient360} shows the average results on the test ODIs measured by the respective metrics on the Salient360$!$ dataset. As can be observed from the table, the proposed model outperforms the other models consistently in terms of NSS and AUC. As for the KL and CC metrics, our proposed model is  inferior to the MV-SalGAN360 \cite{chao2020multi}, but still is very competitive with a KL score of 0.401 and a CC score of 0.658. 
Table~\ref{Table:performanceVR} shows the average results on the test ODIs measured by the respective metrics of these methods on the Saliency in VR dataset. As can be seen from the table, the proposed model achieves a CC score of 0.517, which outperforms the other models consistently. This also demonstrates that a recurrent framework could help achieve better results. Fig.~\ref{comparsion} shows a visual comparison of the results of these test methods on the Salient360$!$ and VR Saliency datasets. As can be observed from the figure, for the other comparative models, some salient areas that exist in the ground truth have not been properly predicted, and some non-salient areas have been wrongly labelled as salient. By contrast, our proposed method can more accurately predict the saliency map with respect to the ground truth density map. Regarding the sizes of the pre-trained models, as shown in Table \ref{Table:ModelSize}, we find that the size of our proposed model is only 2.5 M through experiments. Therefore, it is demonstrated that our proposed model is lightweight and efficient.

\begin{table}[th]
	\centering
	\small
	\caption{Performance comparison after averaging the results of all the test frames on Salient360$!$ dataset \cite{rai2017dataset}.}
	\label{tab:my-table}
	\renewcommand\arraystretch{1.1}
	\begin{tabular}{|c|c|c|c|c|c|}
		\hline Methods & KL$\downarrow$ & CC$\uparrow$ & NSS$\uparrow$ & AUC$\uparrow$ \\
		\hline Maugey \emph{et al.} \cite{maugey2017saliency} & $0.585$ & $0.448$ & $0.506$ & $0.644$ \\
		 SalNet360 \cite{monroy2018salnet360} & $0.458$ & $0.548$ & $0.755$ & $0.701$ \\
		 SalGAN \cite{pan2017salgan} & $1.236$ & $0.452$ & $0.810$ & $0.708$ \\
		 Startsev \emph{et al.} \cite{startsev2018360} & $0.42$ & $0.62$ & $0.81$ & $0.72$ \\
		 GBVS360 \cite{lebreton2018gbvs360} & $0.698$ & $0.527$ & $0.851$ & $0.714$ \\
		 BMS360 \cite{lebreton2018gbvs360} & $0.599$ & $0.554$ & $0.936$ & $0.736$ \\
		 SalGAN\&FSM \cite{de2017look} & $0.896$ & $0.512$ & $0.910$ & $0.723$ \\
		 Zhu \emph{et al.} \cite{zhu2018prediction} & $0.481$ & $0.532$ & $0.918$ & $0.734$ \\
		 Ling \emph{et al.} \cite{ling2018saliency} & $0.477$ & $0.550$ & $0.939$ & $0.736$ \\
		 SalGAN360 \cite{chao2018salgan360} & $0.431$ & $0.659$ & $0.971$ & $0.746$ \\
		 MV-SalGAN360 \cite{chao2020multi} & $\textbf{0.363}$ & $\textbf{0.662}$ & $0.978$ & $0.747$ \\
		 SalBiNet360 \cite{chen2020salbinet360} & $0.402$ & $0.661$ & $0.975$ & $0.746$ \\
		\hline Ours & $0.401$ & $0.658$ & $\textbf{1.09}$ & $\textbf{0.784}$ \\
		\hline
	\end{tabular}
	\label{Table:PerformanceSalient360}
\end{table}

\begin{table}[th]
	\centering
	\small
	\caption{Performance comparison after averaging the results of all the test frames on Saliency in VR dataset \cite{sitzmann2018saliency}.}
	\label{tab:my-table}
	\begin{tabular}{|c|c|}
		\hline Methods & CC$\uparrow$ \\
		\hline Startsev \emph{et al.} \cite{startsev2018360} & $0.431$ \\
		 EB \cite{sitzmann2018saliency} & $0.340$ \\
		 SalNet + EB \cite{sitzmann2018saliency} & $0.470$ \\
		 ML-Net + EB \cite{sitzmann2018saliency} & $0.490$ \\
		 SalNet360 \cite{monroy2018salnet360} & $0.390$ \\
		 SalGAN \cite{pan2017salgan} & $0.361$ \\
		 SalGAN\&FSM \cite{de2017look} & $0.375$ \\
		 SalGAN360 \cite{chao2018salgan360}& $0.488$ \\
		 MV-SalGAN360 \cite{chao2020multi} & $0.507$ \\
		 SalBiNet360 \cite{chen2020salbinet360} & $0.510$ \\
		\hline Ours & $\textbf{0.517}$ \\
		\hline
	\end{tabular}
	\label{Table:performanceVR}
\end{table}

\begin{table}[th]
	\centering
	\small
	\caption{  {Model size comparison of saliency prediction methods.}}
	\label{tab:my-table}
	\begin{tabular}{|c|c|}
		\hline Methods & Model Size(MB) \\
		\hline BMS \cite{zhang2015exploiting} & $27$ \\
		 BMS360 \cite{lebreton2018gbvs360} & $32$ \\
		 GBVS360 \cite{lebreton2018gbvs360} & $34$ \\
		 DVA \cite{8240654} & $96$ \\
		 MLNet \cite{2016A} & $75$ \\
		 SALICON\cite{jiang2015salicon} & $130$ \\
		 SalNet360 \cite{monroy2018salnet360} & $85$ \\
		 SalGAN360 \cite{chao2018salgan360}& $130$ \\
		 MV-SalGAN360 \cite{chao2020multi} & $121$ \\
		\hline Ours & $\textbf{2.5}$ \\
		\hline
	\end{tabular}
	\label{Table:ModelSize}
\end{table}

\section{CONCLUSION}
\label{sec_conclusion}
In this paper, we proposed a novel lightweight saliency prediction moderl for ODI, which is dubbed MRGAN360. We first projected the input ODI into six rectilinear images, then predicted each rectilinear image a saliency map, and finally integrate dthe obtained six saliency maps into a final saliency map for the ODI. For each rectilinear image, we used a multi-stage recurrent generative adversarial network to predict its saliency map progressively, where the recurrent neural network mechanism is used across multiple stages to model the correlations among adjacent stages and a discriminator is used at the end of each stage to supervise the generated saliency map stage by stage. At each prediction stage, the dilated convolutional neural network	was used to acquire a larger receptive field, and the SE blocks were used to assign different alpha-values to different layers according to their channel attentions, to help make more precise saliency prediction. Besides, we shared weights among all the layers to make our model lightweight. Through all the aforementioned operations, the proposed MRGAN360 model is both more accurate and lightweight than the other state-of-the-art methods. In our future work, we plan to improve the model on detecting finer yet more accurate features in high-resolution 360$^{o}$ images. The implementation codes and pre-trained models will be made publicly available.


%

\ifCLASSOPTIONcaptionsoff
  \newpage
\fi



%
\bibliographystyle{IEEEtran}
\bibliography{icme2021template}

%








\end{document}